%% file: main.tex
\documentclass[conference]{IEEEtran}
\IEEEoverridecommandlockouts
\usepackage{cite}
\usepackage{amsmath,amssymb,amsfonts}
\usepackage{algorithmic}
\usepackage{graphicx}
\usepackage{textcomp}
\usepackage{xcolor}
\usepackage{lipsum}
\usepackage{comment}
\usepackage{multirow}
\usepackage{subfigure}
\usepackage{bm}
\usepackage{url}
\usepackage[ruled,vlined]{algorithm2e}
\usepackage{xcolor}
\usepackage{tikz}
\usepackage{textcomp}
\usepackage{hyperref}
\usepackage{fancyhdr}

\fancypagestyle{allpagesstyle}
{
    
    \lhead{\footnotesize THIS PAPER HAS BEEN ACCEPTED IN \underline{\textit{IEEE WCNC 2021}} CONFERENCE AND FINAL VERSION IS SUBMITTED}
    \rhead{\footnotesize \thepage}
    \cfoot{}
}
\newcommand\copyrighttext{
    \footnotesize \textcopyright 2021 IEEE. The final published paper is copyrighted by IEEE and should be cited as: Amr E Hilal, Ismail Arai, Samy El­Tawab, “DataLoc+: A Data Augmentation Technique for MachineLearning in Room­Level Indoor Localization,” IEEE Wireless Communications and Net­working Conference (WCNC), 2021, pp. 1–7. Personal use of this material may be permitted and permission from IEEE must be obtained for all other uses.
}
\newcommand\copyrightnotice{
    \begin{tikzpicture}[remember picture,overlay]
    \node[anchor=south,yshift=10pt] at (current page.south){\fbox{\parbox{\dimexpr\textwidth-\fboxsep-\fboxrule\relax}{\copyrighttext}}};
    \end{tikzpicture}%
}

\def\BibTeX{{\rm B\kern-.05em{\sc i\kern-.025em b}\kern-.08em
    T\kern-.1667em\lower.7ex\hbox{E}\kern-.125emX}}
    
\begin{document}

\title{DataLoc+: A Data Augmentation Technique for Machine Learning in Room-Level Indoor Localization\\
\thanks{\textsuperscript{1} Amr Hilal is also affiliated with the Department of Computer and Systems Engineering, Alexandria University, Egypt.}

\thanks{\textsuperscript{2} Ismail Arai was a visiting scholar at James Madison University, Virginia, USA, while working on parts of this paper.}}

\author{
\IEEEauthorblockN{Amr Hilal \textsuperscript{1}}
\IEEEauthorblockA{Informatics Lab \\ Virginia Tech  \\ Blacksburg, VA, USA \\
ahilal@vt.edu}
\and
\IEEEauthorblockN{Ismail Arai \textsuperscript{2}}
\IEEEauthorblockA{Information Initiative Center\\Nara Institute of Science and Technology\\Ikoma, Nara, Japan\\
ismail@itc.naist.jp}
\and
\IEEEauthorblockN{Samy El-Tawab}
\IEEEauthorblockA{College of Integrated Science and Engineering \\
James Madison University \\
Harrisonburg, VA, USA \\
eltawass@jmu.edu}
}

\maketitle
\copyrightnotice
\thispagestyle{allpagesstyle}

\begin{abstract}
Indoor localization has been a hot area of research over the past two decades. Since its advent, it has been steadily utilizing the emerging technologies to improve accuracy, and machine learning has been at the heart of that. Machine learning has been increasingly used in fingerprint-based indoor localization to replace or emulate the radio map that is used to predict locations given a location signature. The prediction quality of a machine learning model primarily depends on how well the model was trained, which relies on the amount and quality of data used to train it. Data augmentation has been used to improve quality of the trained models by synthetically producing more training data, and several approaches were used in the literature that tackles the problem of lack of training data from different angles. In this paper, we propose DataLoc+, a data augmentation technique for room-level indoor localization that combines different approaches in a simple algorithm. We evaluate the technique by comparing it to the typical direct snapshot approach using data collected from a field experiment conducted in a hospital. Our evaluation shows that the model trained using the proposed technique achieves higher accuracy. We also show that the technique adapts to larger problems using a limited dataset while maintaining high accuracy.

\end{abstract} 


\section{Introduction}\label{sec:intro} 

Indoor Localization has been an active research topic for the past two decades \cite{zafari2019survey,oguntala2018indoor}. The improvement of wireless technology and communication has significantly influenced the research conducted in Indoor Localization, moving from using extra-hardware to more robust and reliable techniques utilizing existing wireless infrastructure. Even though indoor localization research was slowing down, machine learning in communication has raised the topic again to be a hot research topic. With COVID-19 pandemic, indoor localization in a medical environment may have unique importance in identifying or tracing equipment or certain individuals at a specific time at a Medical Facility.

\begin{figure}[htbp] 
    \centering
    \includegraphics[width=\linewidth,clip]{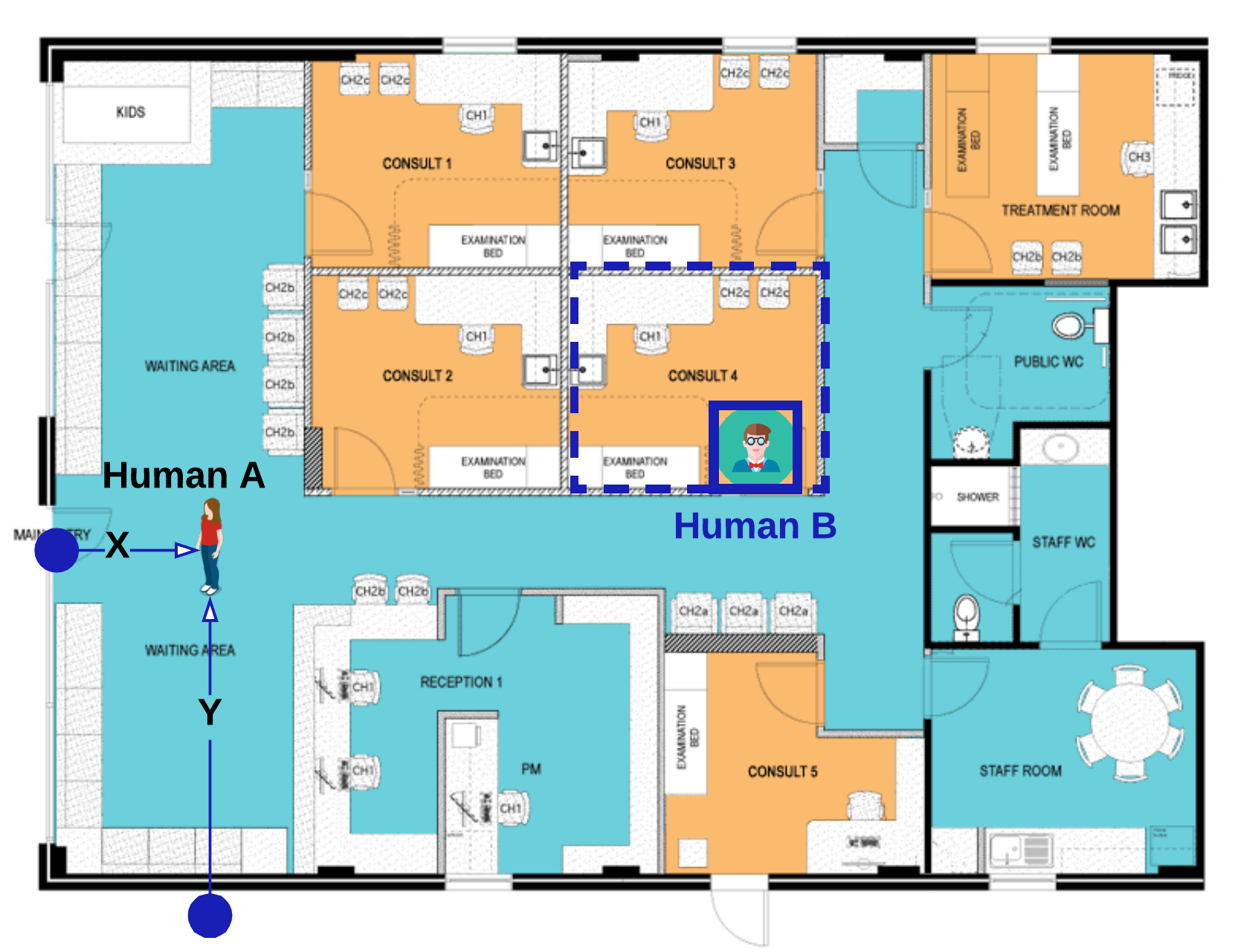}
    \caption{Precise Positioning vs. Room-level Localization}
    \label{fig:local}
\end{figure}

An indoor localization system can be classified into one of two types depending on the applications (e.g., commercial, individual tracing, military, industrial inventory tracking, and retails); coordinate-level and room-level positioning. Researchers are trying to answer the question that mainly depends on the application type. Some applications may require detection of the target object direction (e.g., indoor navigation in a museum)\cite{alletto2015indoor}. In Fig.~\ref{fig:local}, the system is interested to know the precise position of Human A (x, y coordination and possibly the direction). In contrast, the system is interested in room-level localization for Human B (e.g., at Consulate Room no.4 is a sufficient answer).

Data is essential for any machine learning process. Without it, a model cannot be trained to perform the function it is intended to do. Both quantity and quality aspects are necessary to produce a model well-imitating to a phenomenon or an activity. Fingerprint-based indoor localization techniques have majorly relied on signal strength of surrounding access points as a signature to train machine learning models to identify indoor locations \cite{chen2016robust}. However, collecting the needed data from different indoor locations is usually a daunting task. This data is usually collected using internal system tools that produce a snapshot representing signal strength of the surrounding access points. After collection, snapshots are formatted into data samples to train a machine learning model on the signal strength signature at each particular location. Typically, more data samples, among other considerations, help improve machine learning accuracy and reduce overfitting; however, the lack of real-life data can be a challenge \cite{wang2018wifi,guo2018indoor}.

To overcome the training data problem, data augmentation has been used to synthetically produce more training data \cite{sinha2019data, rizk2018cellindeep}. Besides quantity, data augmentation helps researchers introduce data from different scenarios that is otherwise hard to produce in real experiments. In this paper, we introduce DataLoc+, a data augmentation technique that produces context-related data for room-level indoor localization. The technique uses a simple algorithm to produce location signatures for different scenarios of loss or variations in the signal strength received from the surrounding network devices. These scenarios used to be produced in the literature using separate techniques such as random dropping of network devices and distribution fitting of signal strength \cite{rizk2019effectiveness}. We evaluate the model using the data we collected from our experiment at Sentara hospital (located at Harrisonburg, VA, USA) by comparing it to the more conventional direct snapshot technique. We also show that the proposed technique can adapt to larger classification problems with the same amount of real-data collected from the environment. 

The remaining of the paper is organized as follows, in section~\ref{sec:related} we shed light on related work on the topic of indoor localization, and the integration of machine learning techniques into indoor localization. Section~\ref{sec:scaled} provides details of the technique proposed in this paper. In section~\ref{sec:method}, we describe our research methodology, including data collection and the evaluation. A summary and discussion of the results is provided in section~\ref{sec:analysis}. We conclude the paper and highlight our future directions in section~\ref{sec:final}.

\section{Related Work}\label{sec:related}

The localization problem has been extensively studied in the literature both indoor and outdoor \cite{vo2015survey,cheema2018indoor,salman2018indoor}. Several technologies also have been used as a source of location identification like Bluetooth \cite{teran2017iot}, WiFi \cite{wang2018wifi}, visual landmarking \cite{li2017visual}, and even device-free \cite{youssef2007challenges}. Recently, indoor localization has gained momentum for improvement using machine learning techniques \cite{rizk2018cellindeep,abbas2019wideep}. Salamah et al. proposed using machine learning to enhance the accuracy of WiFi-based indoor localization while reducing the required computational cost and time \cite{salamah2016enhanced}. Farid et al. \cite{farid2016hybrid} proposed a hybrid technique for the implementation of indoor localization. They adopted fingerprinting in both WiFi and Wireless Sensor Networks (WSNs), where their model exploits machine learning for position calculation using Artificial Neural Networks (ANN). 

Besides neural networks, some simple machine learning techniques were studied and found effective. Cadleroni et al. \cite{calderoni2015indoor} used Random Forest classifiers in a hospital environment for indoor localization. Their system was based on the Radio-Frequency Identification (RFID) technology and a hierarchical structure of classifiers. AlHajri at al. \cite{alhajri2019indoor} presented a two-stage machine learning approach to achieve high localization accuracy by harnessing the richness of the RF features, e.g., Received Signal Strength (RSS) and Channel Transfer Function (CTF), in an indoor environment. In the first stage, they used machine learning to identify the type of indoor environment based on previously generated indoor RF profiles. In the second, machine learning was used to pick the best combination of RF features that yield the highest localization accuracy. They showed that the simple $k$-NN classifier was fast and effective in achieving high accuracy, and concluded that it is a successful candidate for real-time IoT deployment scenarios.

Data augmentation has been a key tool in machine learning-based localization approaches to overcome the lack of training data \cite{xiang2019robust, rizk2020gain}. Sinha et al. \cite{sinha2019data} proposed two data augmentation schemes for deep learning architectures used in fingerprint-based indoor localization using RSS values. Their schemes aimed to diversify the training RSS values by constructing new data samples based aggregate statistics calculated using the existing samples. Rizk et al. \cite{rizk2019effectiveness} studied the effectiveness of data augmentation in cellular-based localization using deep learning. Their methodology focused on cellular-based localization using deep-learning techniques. They proposed different techniques and showed a performance improvement in both indoor and outdoor scenarios. They also utilized data augmentation in their cellular-based localization technique in \cite{rizk2018cellindeep}.

\section {Description of The Proposed Technique} \label{sec:scaled}

In fingerprint-based indoor localization, building a radio map in the offline phase is a prerequisite to recognizing locations in the online phase. In this map, a location is identified by the signature of radio characteristics, most commonly signal strength, received from surrounding wireless devices running Bluetooth, WiFi, or Cellular technology. Machine learning has been increasingly used to replace these fingerprint maps by training a model to identify a location using its wireless location signature.

When relying on WiFi devices for indoor localization, the dominant way to collect data to train a machine learning model is to record snapshots of the signal strength received from the surrounding WiFi access points using a system tool on a laptop or mobile device. Due to environmental obstacles that lead to fading and shadowing effects, signal strength recorded at a particular location can suffer from large variations reflecting in the data samples used to train the model. The environmental factors can also lead to a temporary loss of signals received from some network devices. The model needs to be trained on such scenarios to be able to recognize them in reality. 

The proposed technique in this paper is inspired by the dropout technique used to prevent overfitting in neural networks \cite{srivastava2014dropout}. It seeks to inject the different scenarios the signal strength may suffer from into the produced data by variantly dropping frames from a received stream of broadcast beacons. This is done by taking a step back in how signal strength snapshots are usually collected. Instead of relying on the values reported by system tools, we track the incoming stream of broadcast beacon frames received from the surrounding network devices and use a variable-size representative window to produce snapshots. A snapshot is generated using a portion of the most recent window of beacon frames, where the frames are randomly shuffled before a portion (e.g. 80\%) is used. An access point is represented in a snapshot by averaging the signal strength associated with all beacon frames received from that particular access point. 

The portion dropped every time is what injects the signal strength loss and variation effect into the data in an abundance of forms. The portion used to generate the snapshots can be scaled up or down and move along different ranges depending on the nature of the environment. This range of portions is what controls the variability in the produced snapshots. Low portions will introduce more losses in detected network devices and more variability in the detected signal strength because less number of values are averaged. High portions, on the other hand, tend to inject a more complete set of network devices and less variable values of the signal strength. In the online phase, a mobile device roaming an indoor environment will use the same windowing techniques described above to produce its snapshot except that no portions will be dropped. Algorithm~\ref{alg:DataLoc} shows how snapshots are generated using the DataLoc+ technique.

\begin{algorithm}[htbp]
    \label{alg:DataLoc}
    \SetAlgoLined
    \SetKwInOut{Input}{input}\SetKwInOut{Output}{output}
    \Input{$startPercentage, endPercentage, step, reps$}
    \Output{snapshots at location $X$}
    \BlankLine  
	$beacons \gets$ \text{received beacon frames at location $X$}\\
	$snapshots \gets$ \text{empty array}\\
	\BlankLine
	\For{$portion \gets startPercentage$ \KwTo $endPercentage$}{
		\For{$i \gets 1$ \KwTo $reps$} {
			randomize order in beacons \\
			$chunk \gets$ \text{extract portion from beacons}\\
			$signature \gets$ \text{average signal strength of}\\
			\hspace{1.95cm} network devices in chunk\\
			$snapshots \gets$ \text{add $signature$}\\
		}
		$portion \gets portion + step$
		\BlankLine
	}
    \caption{DataLoc+}
\end{algorithm}

\section{Methodology} \label{sec:method}

\subsection{Data Collection}
This research work is conducted as part of a 4-VA funded project that aims to improve accuracy for room-level indoor localization for the purpose of tracking patients and equipment inside medical facilities. Therefore, we rely in this study on the data we collected from a field experiment in Sentara hospital (in Harrisonburg, Virginia) in February 2020. We collected data in three types of spaces in the hospital; patient and office rooms, operation rooms, and encompassing corridors. We were interested in all types of network devices that are either access points or acting as access points (e.g., printers, surveillance cameras, TVs, etc.), which can be increasingly observed in indoor environments \cite{hilal2019exploring}.

\begin{figure}[htbp] 
    \centering
    \includegraphics[width=0.45 \textwidth]{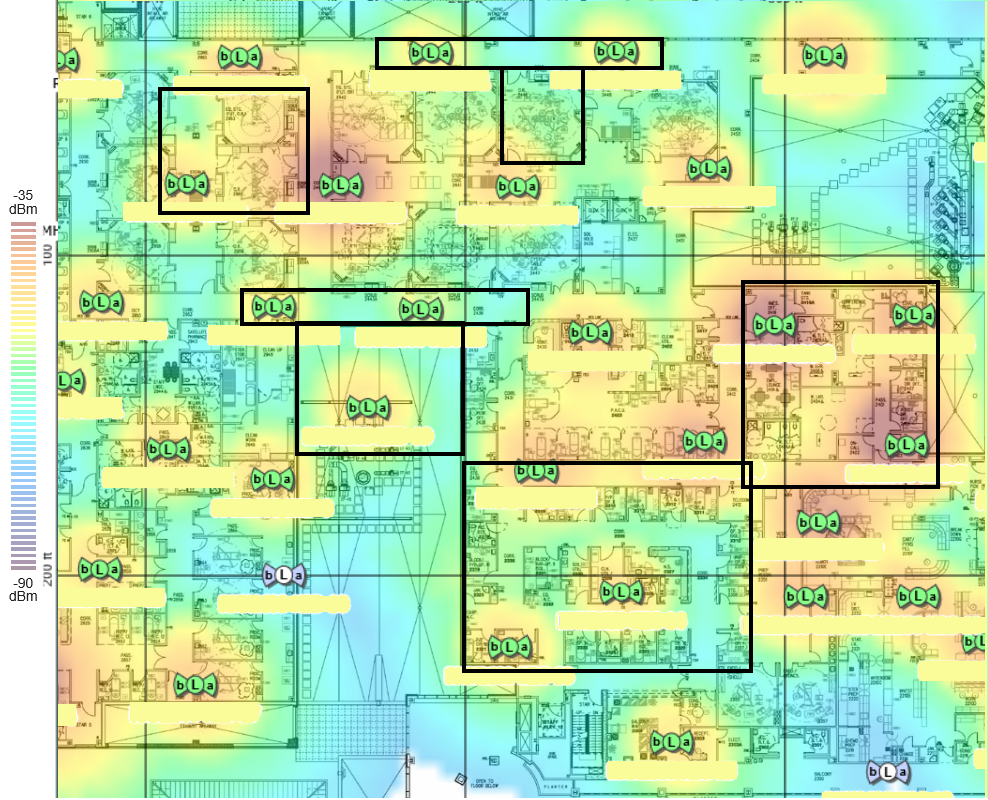}
    \caption{Floorplan at Sentara Hospital}
    \label{fig:floorplan}
\end{figure}

Fig.~\ref{fig:floorplan} shows the floorplan of the investigated floor in the hospital where the areas we were permitted to enter are marked up. To better test the accuracy of our trained models, we did our best to collect data from adjacent rooms; however, that was not always possible. For example, we were permitted to patient rooms only if they were unoccupied. Similarly, the staff were very cooperative and allowed us to collect data from their offices and common areas. Still, some offices were busy in the limited time window we were in the hospital. We were also permitted to enter 4 operation rooms where we were asked to wear special protective garment. The operation rooms were much bigger than typical patient rooms; therefore, we collected data from more positions inside them. 

To study the operability and effectiveness of DataLoc+, we use it to produce training data and compare the accuracy of the resulting model to that of a model that has been trained using the more conventional snapshot collection approach. To achieve this comparison, we collected data for model training from multiple positions at different rooms in two modes:

\begin{itemize}
    \item \textbf{Snapshot Mode:} The conventional way of collecting snapshots. Data about surrounding network devices, including received signal strength, is captured separately and repeatedly using laptop or mobile device system tools without using data augmentation. The \texttt{airport} command was used to collect these snapshots from a MacBook Air laptop. 
    \item \textbf{Stream Mode:} Streams of broadcast beacon frames are collected from surrounding network devices across different channels on the 2.4 GHz and 5 GHz bands. Strength of the signal received from a network device is included in its corresponding beacon frame. The \texttt{tcpdump} command was used with qualifying filters to capture broadcast beacon frames while running a channel hopping script to scan the channels across the two bands. Data collection was done on a MacBook Air as well. A capturing session lasted from 40 to 80 seconds at each position capturing 1000 to 2500 broadcast beacon frames. The number of detected network devices also affects the number of collected beacon frames.
\end{itemize}

Data was collected from 2 to 4 positions in office/patient rooms, 5 to 9 positions in operations rooms, in addition to positions along the corridors. In both modes, for each detected network device, we extracted its BSSID (MAC address), SSID (network name), and detected signal strength. The collected data is used to build a feature file that is then used to train a machine learning model. In the feature file, network devices, mainly access points, are considered the features (columns), and snapshots of corresponding signal strength are considered the data samples (rows). A label marking the space the snapshot was taken from is associated with each data sample (e.g., room1, room2, etc.). 

\bgroup
\def\arraystretch{1.3}
\begin{table}[htbp]
\caption{Experimental Setup}
\centering
\begin{tabular}{|c|c|c|}
\hline
\textbf{Paramter} & \textbf{Stream mode} & \textbf{Snapshot mode}  \\ \hline
\textbf{Device} & \multicolumn{2}{c|}{Macbook air} \\ \hline
\textbf{WiFi bands} & \multicolumn{2}{c|}{2.4 and 5 GHz} \\ \hline
\textbf{Locations} & \multicolumn{2}{c|}{35 office, patient, operations and hallways} \\ \hline
\textbf{Position per room} & \multicolumn{2}{c|}{2-4 in office/patient, 5-9 in operations} \\ \hline
\textbf{Command/tool} & tcpdump & airport -s \\ \hline
\textbf{Detected devices} & 280 & 260 \\ \hline
\textbf{Capturing session length} & 40-80 sec & N/A \\ \hline
\textbf{Frames per position} & 1000-2500 & N/A \\ \hline
\end{tabular}
\label{exp_steup}
\end{table}

Building the feature file in the snapshot mode is done by parsing all snapshots and extracting access point information, including signal strength and identifying MAC address. Then, for each data sample, the signal strength for undetected MAC addresses are given a -100 dBm to represent undetectability. Same steps apply to the stream mode except that each data sample represents one beacon frame, so only the signal strength value associated with the network device that originated that frame exists in this snapshot. DataLoc+ is then applied to these single-value snapshots as described in the previous section to produce the complete snapshots. These snapshots constitute the final feature file that is fed to the machine learning model for training. A total of 260 network devices were detected in the snapshot mode, while 280 were detected in the stream mode (excluding hidden access points). Table \ref{exp_steup} summarizes the experiment parameters in both modes.

\subsection{Evaluation}

\begin{figure}[htbp] 
     \centering
     \includegraphics[width=0.4\textwidth]{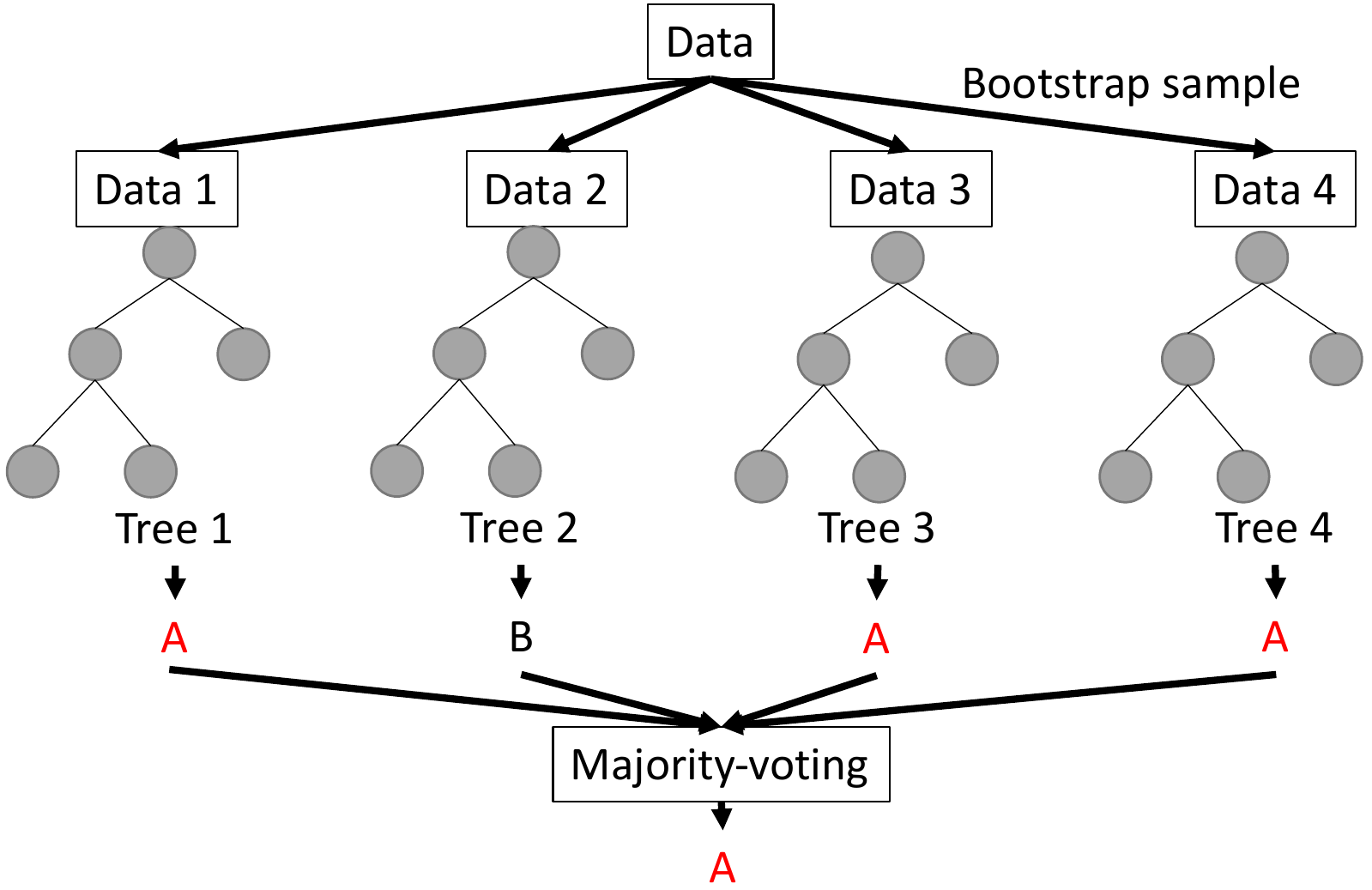}
     \caption{Overview of Random Forest}
     \label{fig:rf-overview}
\end{figure}

Room-level indoor localization is a classification problem where rooms are the sought classes. Inspired by \cite{calderoni2015indoor}, we used a Random Forest classifier to evaluate our proposed data augmentation technique. Random Forest is an ensemble model of decision trees where multiple models are trained with different portions of the data and a prediction is made through a majority voting among the participating models as illustrated in Fig.~\Ref{fig:rf-overview}. Random Forest is more robust for overfitting than a sole decision tree which tends to overfit on training data with a large number of features. Random Forest avoids overfitting by randomizing data and features. In the former, each decision tree in the Random Forest is built from a sample drawn with replacement from the training set (AKA a bootstrap sample). In the latter, each decision tree has its own split features. Though each tree has less accuracy than one big decision tree, the combined forest outputs higher accuracy than a big decision tree by taking an average of the trees and decreasing its variance.

Random Forest has three main hyper-parameters, Max Features, Max Depth, and Number of Estimators. Max Features is the maximum number of features applying to each decision tree. Usually, the square root of the total number of features is the chosen default value. Max Depth is a maximum depth of a tree. A deeper tree outputs higher accuracy at the training phase, though a moderate depth is necessary to avoid overfitting. Number of Estimators is the number of trees constituting the forest, and a moderate depth is also needed to avoid overfitting. We used the scikit-learn framework \cite{scikit-learn} in our analysis and focused mainly on the Max Depth and Number of Estimators in our evaluation.

\section{Results}\label{sec:analysis}

\subsection{Effect of sliding portions}
To show the effect of sliding portions on the variability of the produced snapshots, we look at the produced signal strength and percentage of access  points included, both across different portions. Fig.~\ref{fig:variability-SS} shows the produced signal strength by DataLoc+ for an example access point at a particular position as portions increased (others exhibited the same behavior), 20 \textit{reps} per portion. The minimum and maximum signal strength detected from the access point examined in the figure were -78 and -72 dBm respectively. The figure shows that signal strength values of larger variability are produced at low portions because less values are averaged, while the value centers around the mid-range as the portion approaches 1. The portion parameter, therefore, can be used to produce signal strength values that lie within the original detected range.

Fig.~\ref{fig:variability-APs} shows the other aspect of variability snapshot data need to mimic, that is the dropped access points. 105 APs were originally detected from 1863 beacon frames at the position examined in the figure. The figure shows that the percentage of included access points increased as the portion increased. However, the percentage value will differ if the number of collected frames increased significantly. 

\begin{figure}[htbp] 
     \centering
     \includegraphics[width=0.4\textwidth]{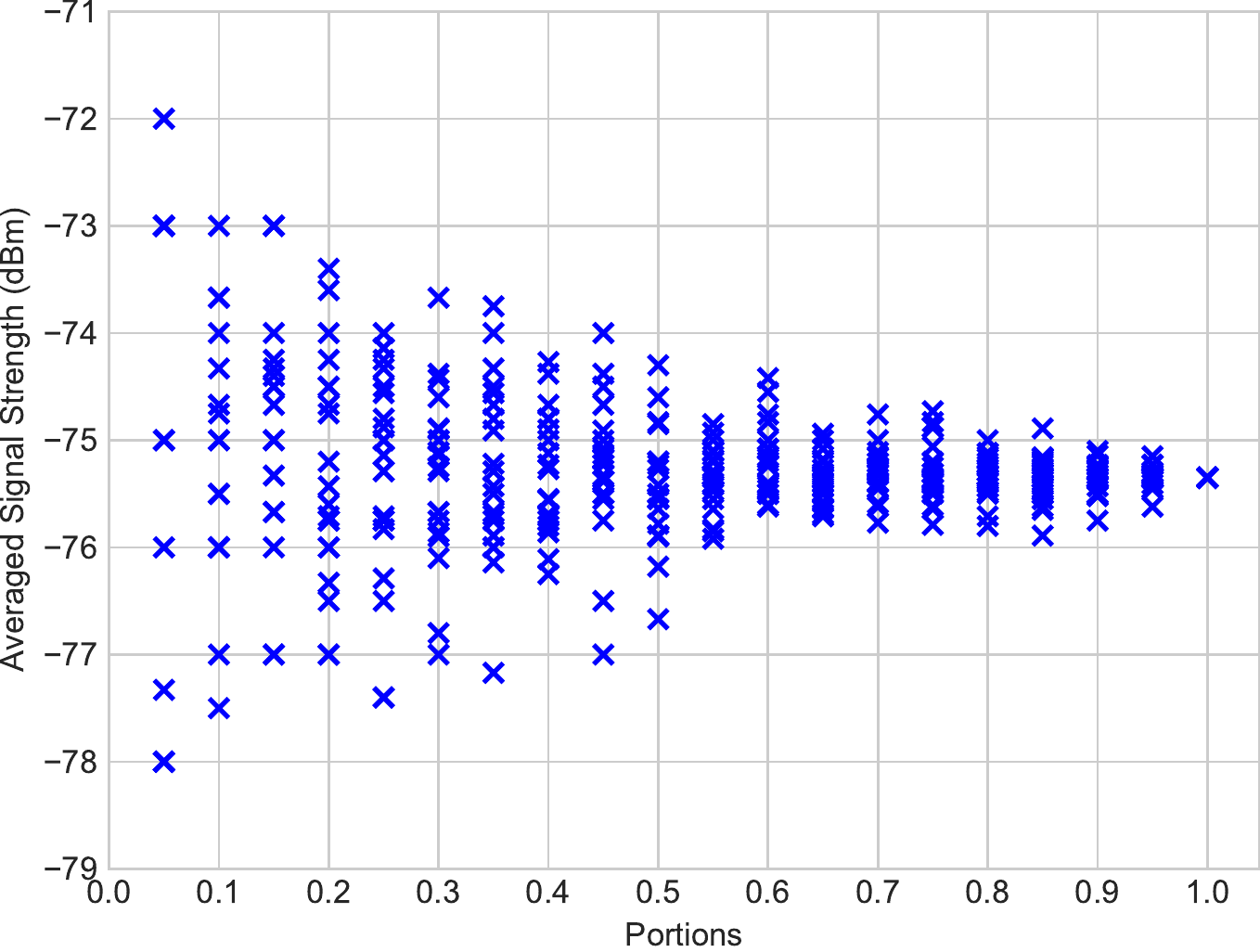}
     \caption{Signal strength at a particular access point produced at different portion values.}
     \label{fig:variability-SS}
\end{figure}

\begin{figure}[htbp] 
     \centering
     \includegraphics[width=0.4\textwidth]{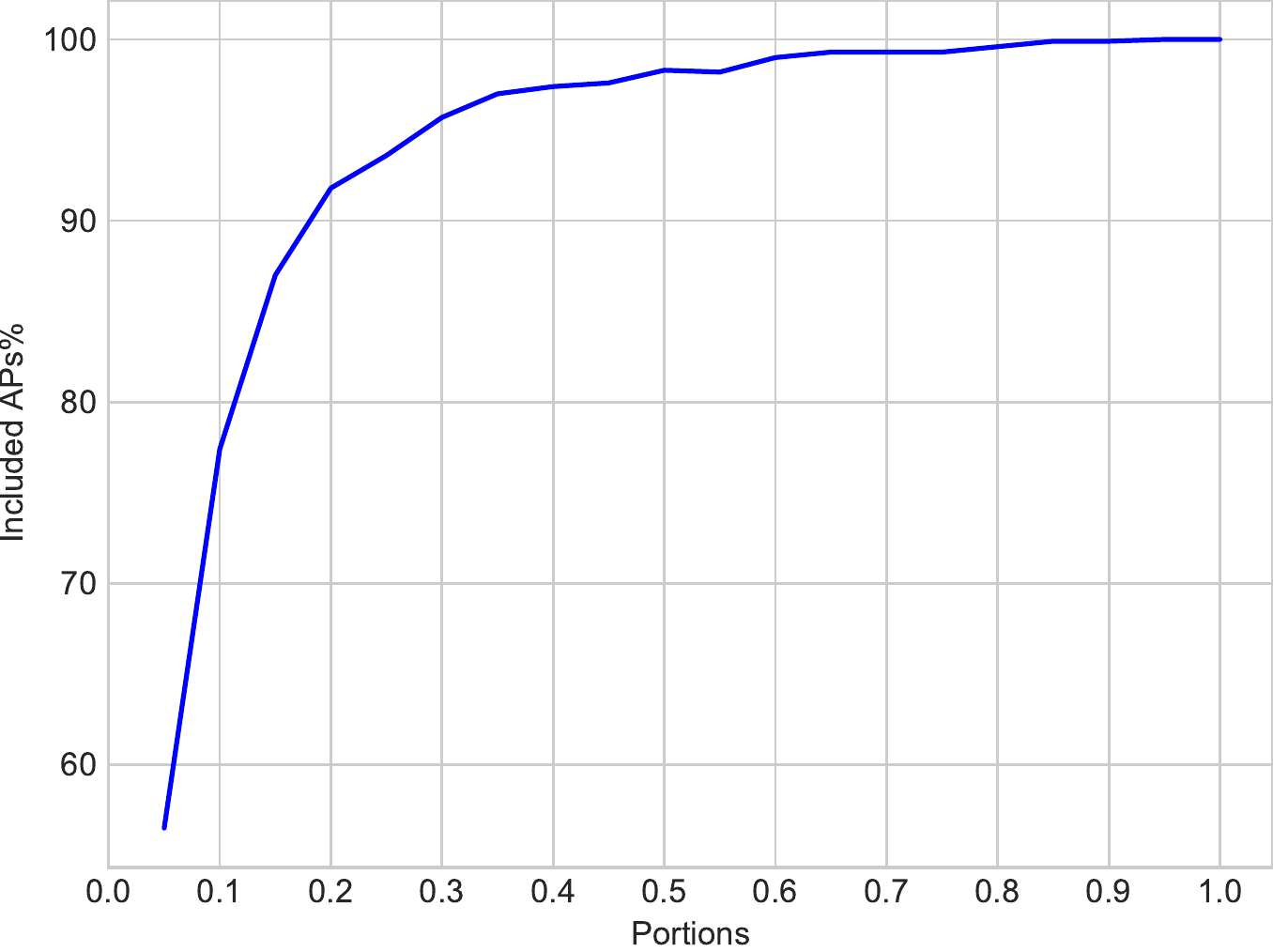}
     \caption{Percentage of included access points as portion increases.}
     \label{fig:variability-APs}
\end{figure}

\begin{figure*}[t]
    \centering
    \subfigure[Snapshot technique]{
        \includegraphics[width=0.48\textwidth]{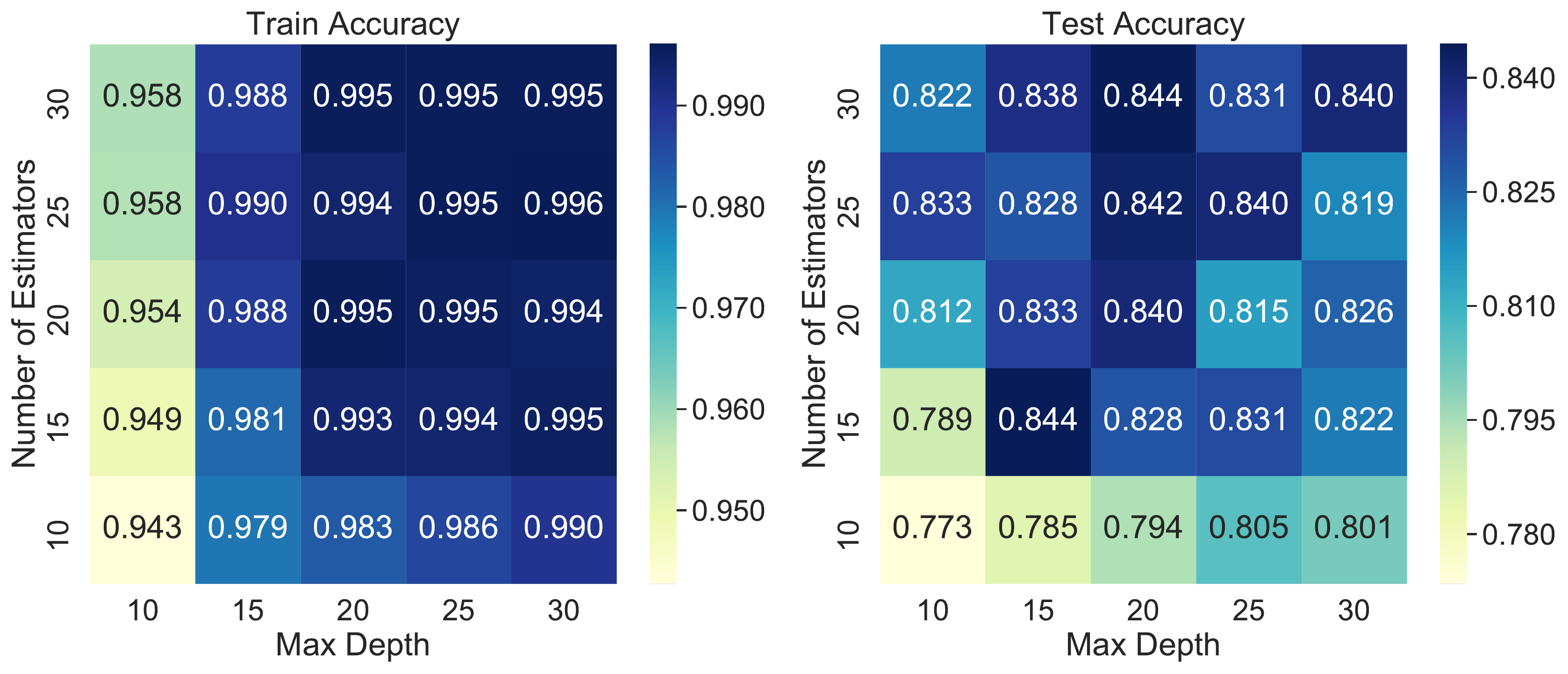}
        \label{fig:snapshot-technique}
    }
    \subfigure[DataLoc+ at (0.2, 1.0, 0.2, 1)]{
        \includegraphics[width=0.48\textwidth]{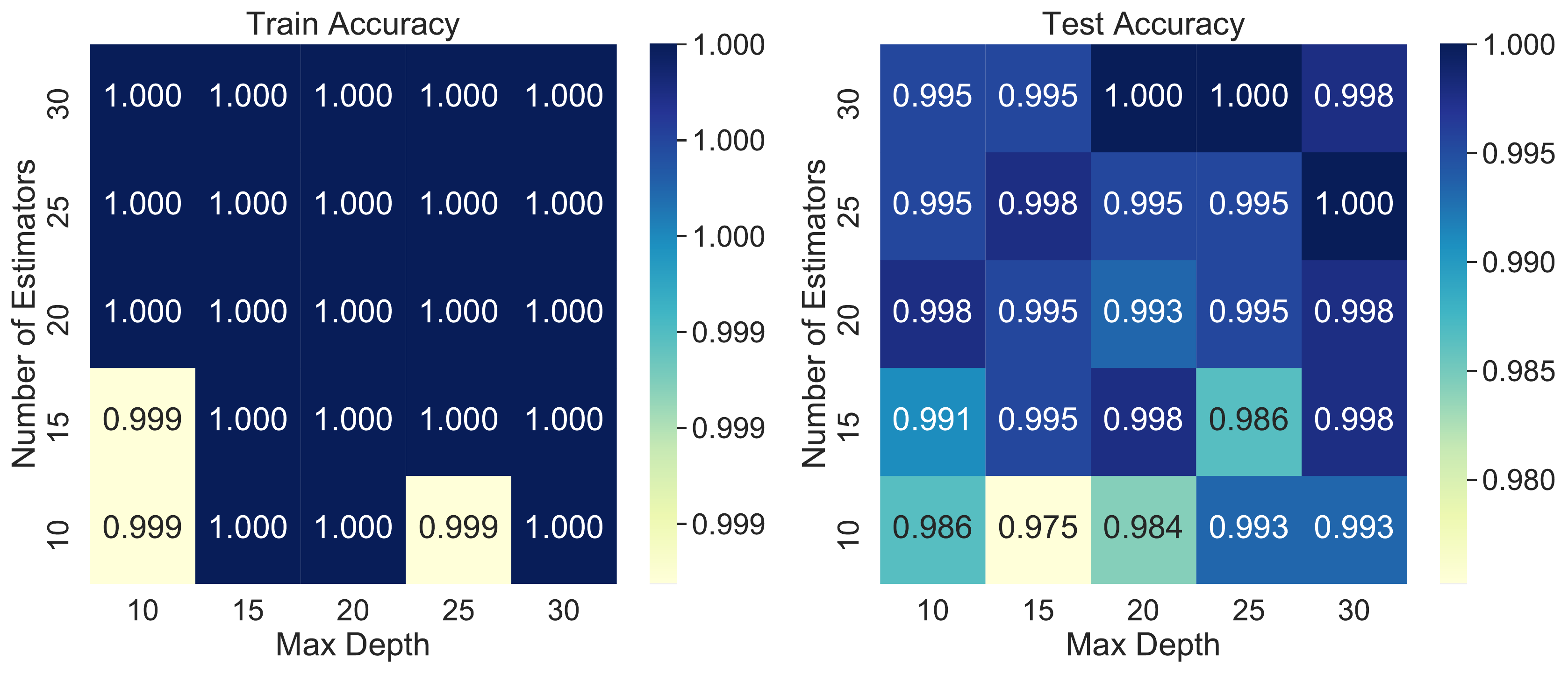}
        \label{fig:snapshor-vs-model-b}
    }
    \par\medskip
    \subfigure[DataLoc+ at (0.5, 1.0, 0.125, 1)]{
        \includegraphics[width=0.48\textwidth]{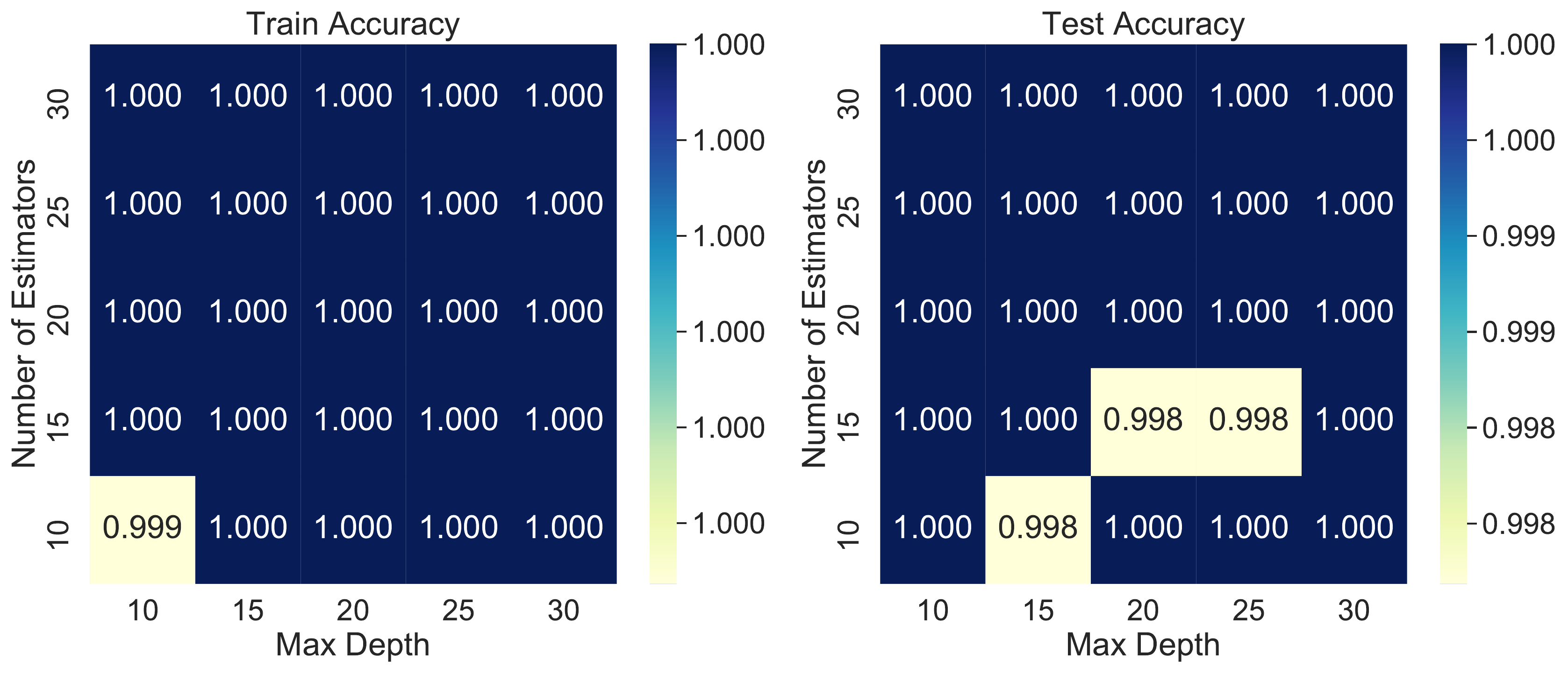}
    }
    \subfigure[DataLoc+ at (0.8, 1.0, 0.05, 1)]{
        \includegraphics[width=0.48\textwidth]{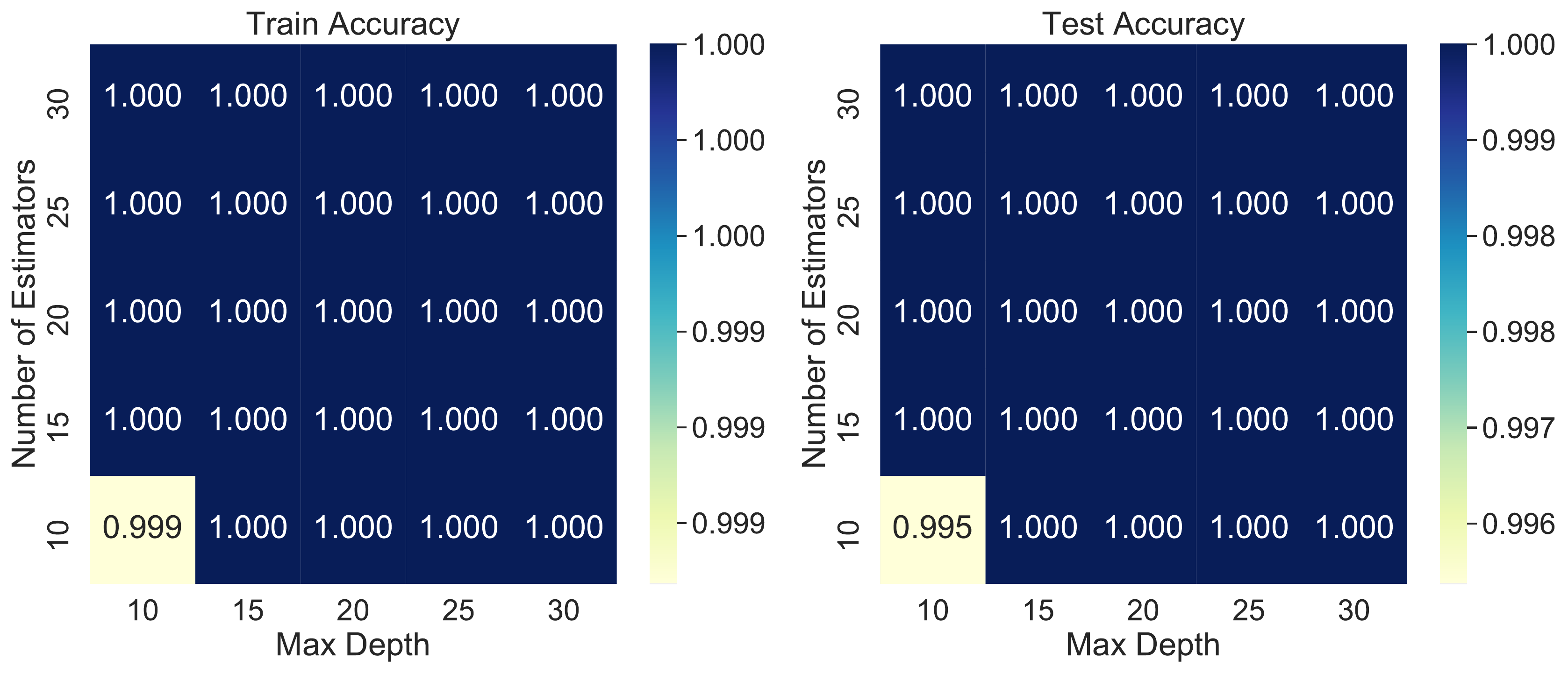}
        \label{fig:snapshor-vs-model-d}
    }
    \caption{Grid view of train and test localization accuracy of the snapshot technique vs. different range of portions of DataLoc+ using 450 data samples}
    \label{fig:snapshot-vs-model}
\end{figure*}

\subsection{Comparison to the conventional snapshot technique}
To conduct this comparison, we train two models on the data generated in the snapshot and stream modes. We study accuracy changes with the two modes across a grid of hyper-parameters of the Random Forest classifier, namely \textit{maximum depth} and \textit{number of estimators}. Since the number of data samples in the snapshot technique is bound by the number of readings collected, we adjusted the proposed data augmentation technique to produce an equivalent number of data samples. We tried 3 ranges of portions across the data and expressed a range of portions in this notation \textit{(start percentage, end percentage, step, reps)}. A range of portions \textit{(0.4, 1.0, 0.2, 5)} means randomly shuffling the data and taking portions starting from 40\%, repeating 5 times with shuffling, then moving the percentage up 20\%, until reaching 100\%. 

Fig.~\ref{fig:snapshot-vs-model} shows a comparison of train and test accuracy of the two techniques plotted over a grid of different values of the Random Forest hyper-parameters \textit{maximum depth} and \textit{number of estimators} ranging from 10 to 30 for both in the form of a heatmap. DataLoc+ is run at 3 portion ranges \textit{(0.2, 1.0, 0.2, 1), (0.5, 1.0, 0.125,1), (0.8, 1.0, 0.05, 1)}. The number of data samples used for both experiments is around 450. The number of spaces examined is 35 (classification labels), including office/patient rooms, operation rooms, and different segments of the corridors. In machine learning, grid search is performed as part of the Hyper-Parameter Optimization (HPO) process, where the best combination of classifier hyper-parameters is sought for achieving the best accuracy before applying to test data. We use grid search in our context primarily to compare train and test accuracy of an escalating range of Random Forest hyper-parameters across the two techniques. By traversing the space of hyper-parameters, model accuracy improves then may deteriorate due to overfitting. Therefore, the best combination of parameters could be found in the middle of the space, as can be seen in Fig.~\ref{fig:snapshot-technique}. Observing a sustained small difference between the corresponding train and test accuracy rules out the possibility of overfitting, while observing a sustained high accuracy value rules out the possibility of underfitting. Fig.~\ref{fig:snapshot-vs-model} shows that the best test accuracy observed in the conventional snapshot technique is 84.4\% at 20 and 30 Max Depth and Number of Estimators respectively, while it approached 100\% in most of the corresponding DataLoc+ settings. That is a 15\% accuracy improvement for a model trained using data produced by DataLoc+ using the same amount of data, classes, and hyper-parameter search space. The less the range of portions is extended (Fig.~\ref{fig:snapshor-vs-model-d} is less extended than Fig.~\ref{fig:snapshor-vs-model-b}), the more consistent snapshots are produced because each snapshot is produced using more beacon frames yielding better accuracy compared models trained with the same amount of data generated in the conventional snapshot mode.

\subsection{Adaptability to larger problems}

As pointed out before, data is the fuel to train a machine learning model, and both quantity and quality are equally important. Training a model for a larger problem (more classes to distinguish) requires more data. We study here how DataLoc+ can expand a limited amount of data to train a larger problem. To do this, we needed to collect data for a larger space with more distinguishable separable rooms. However, that was not possible when we worked on this study due to the lockdown caused by the COVID-19 pandemic. To overcome this problem, we subdivided each distinguishable zone (all types of spaces described above) into multiple sub-zones representing the fixed positions at which we collected data within the original zone. With an average of 3 positions per zone, we could almost triple the problem size in terms of distinguishable classes. Each sub-zone is less in space and amount of training data available. We then used our proposed data augmentation technique to produce more training data for each sub-zone.

Fig.~\ref{fig:model-increased-portions} shows the train and test accuracy of a model trained with data produced by DataLoc+ over different ranges of portions and fixed space of Random Forest hyper-parameters. As the portion settings are adjusted to produce more data, the model accuracy improves. Although it is usually expected in machine learning to have accuracy improvement as training data increases, other factors can impact that. Like in Fig.~\ref{fig:snapshot-vs-model}, the numbers rule out the possibilities of overfitting and underfitting. However, we wanted to show from Fig.~\ref{fig:model-increased-portions} that using the same pool of collected real-data, we could generate more quality data to better train a machine learning model for improved indoor localization accuracy. The limit to which data generation can extend without impacting data quality is subject to a further study.

\begin{figure*}[htbp]
    \centering
    \subfigure[Sub-zones at (0.2, 1.0, 0.2, 1), 450 data sample]{
        \includegraphics[width=0.48\textwidth]{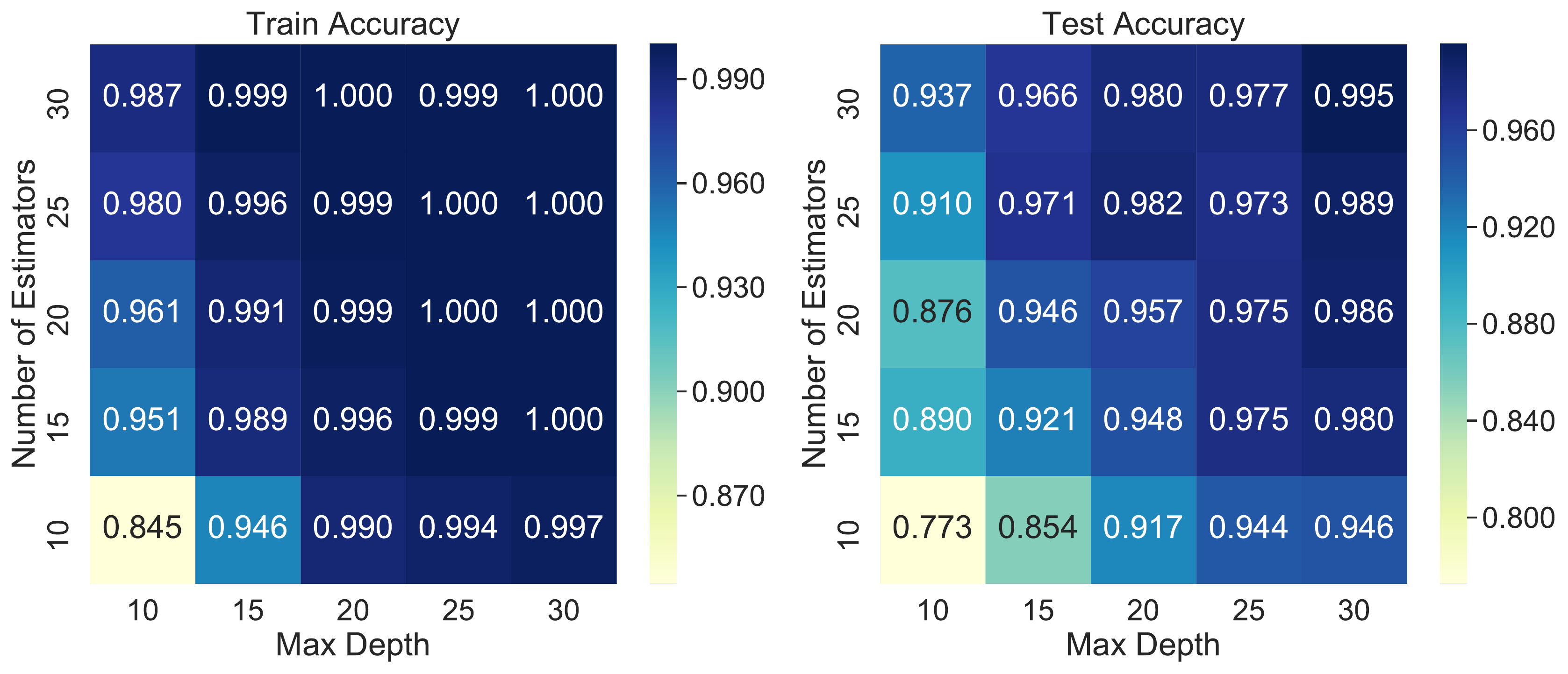}
    }
    \subfigure[Sub-zones at (0.2, 1.0, 0.1, 1), 800 data sample]{
        \includegraphics[width=0.48\textwidth]{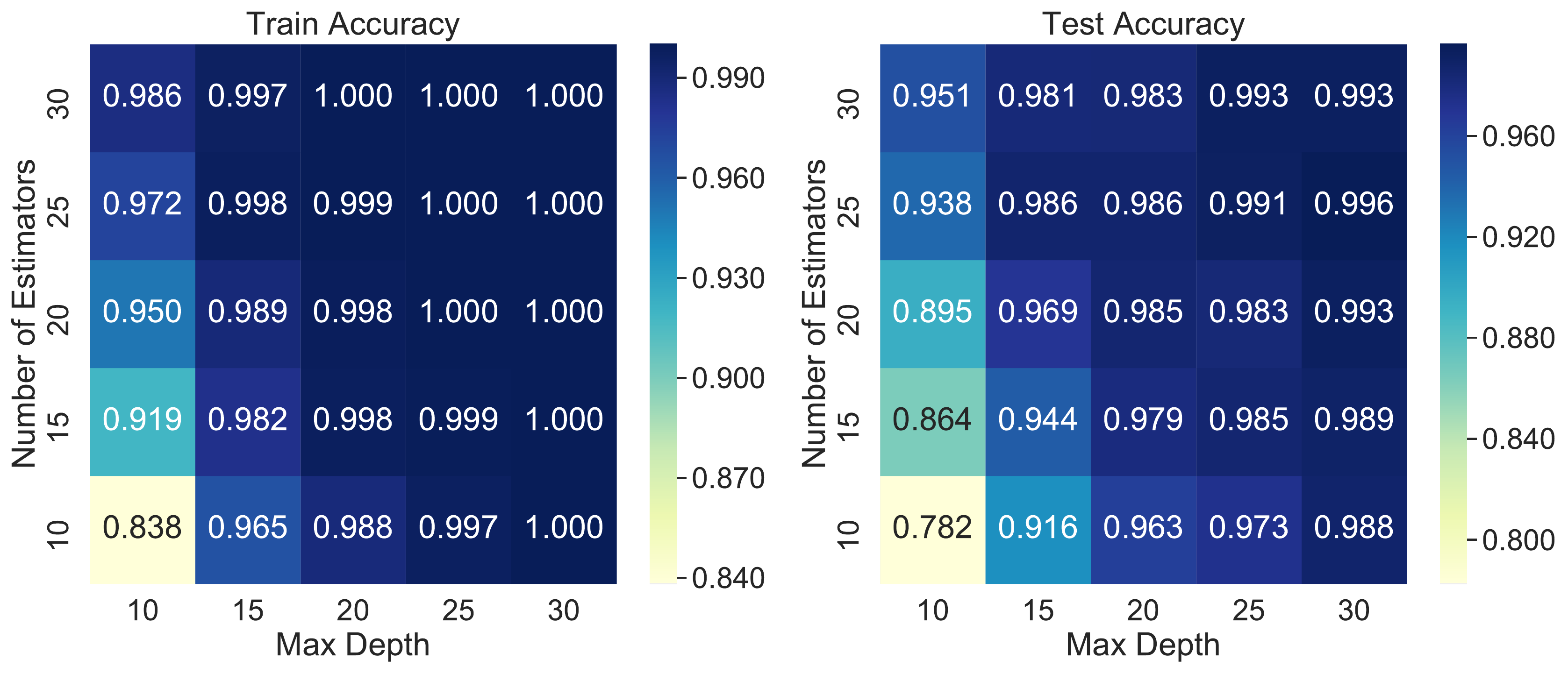}
    }
    \par\medskip
    \subfigure[Sub-zones at (0.2, 1.0, 0.05, 1), 1600 data sample]{
        \includegraphics[width=0.48\textwidth]{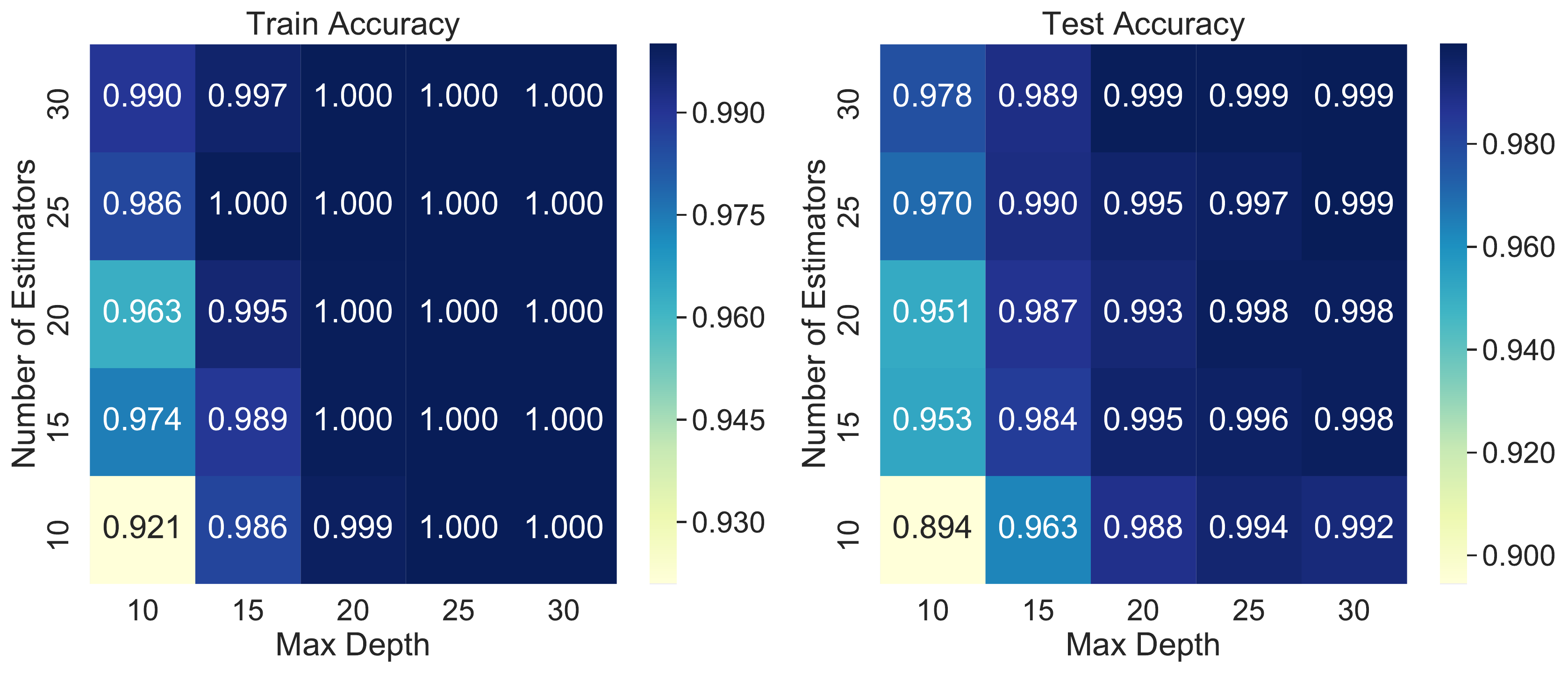}
    }
    \subfigure[Sub-zones at (0.2, 1.0, 0.05, 2), 3200 data sample]{
        \includegraphics[width=0.48\textwidth]{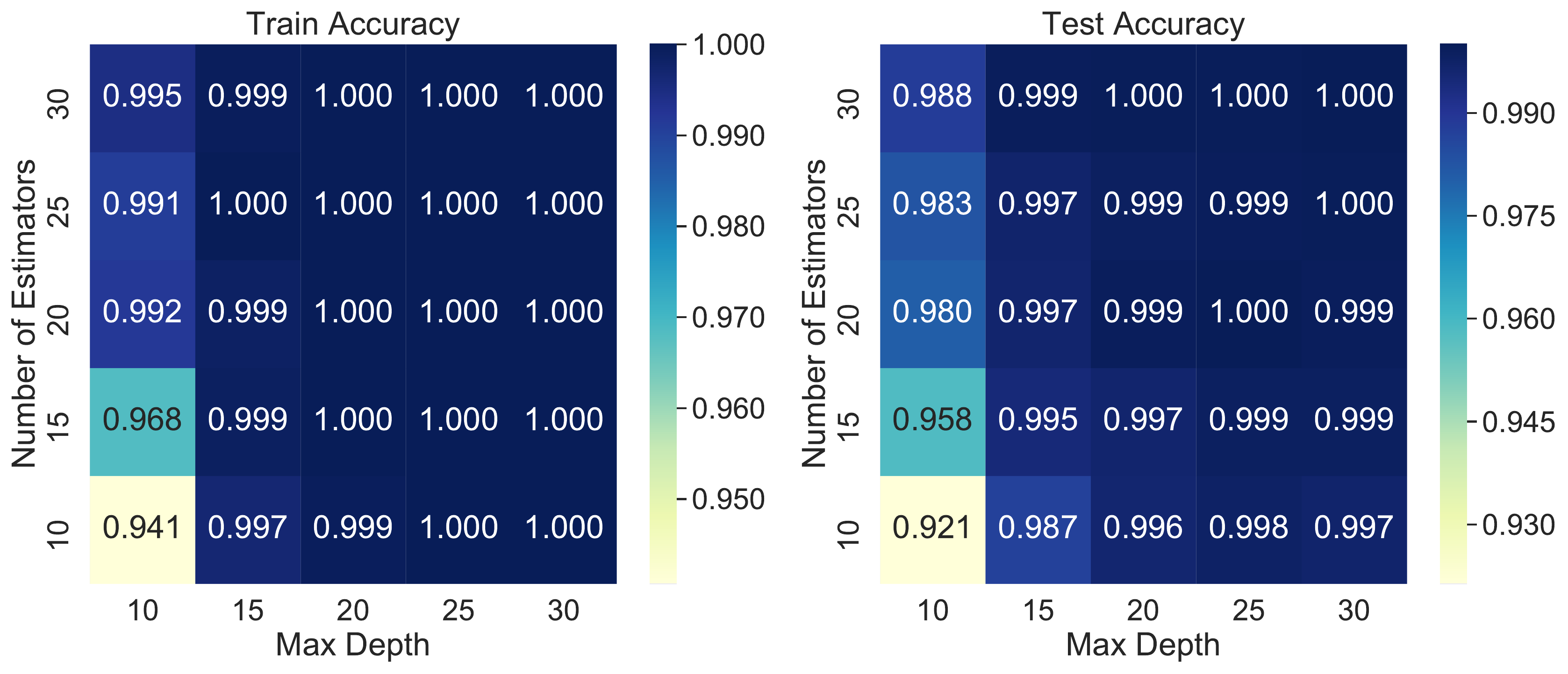}
    }
    \caption{Grid view of train and test localization accuracy of different range of portions of DataLoc+ for sub-zone locations}
    \label{fig:model-increased-portions}
\end{figure*}

\section{Conclusions and Future Directions}\label{sec:final}

The work in this paper focuses on the problem of data availability for machine learning in room-level indoor localization. Particularly, the more data representing an indoor environment is used, the better localization accuracy can be attained. We proposed DataLoc+, a data augmentation technique that aims to extend generation of quality data for the purpose of improving model accuracy by combining different approaches in one algorithm. DataLoc+ achieves this goal by tracking the stream of broadcast beacon frames carrying the signal strength of surrounding network devices. Snapshots are then generated by averaging the signal strength over variable portions of the most recent window of beacon frames. We showed that the proposed technique produces 15\% higher accuracy than the conventional snapshot technique. We also showed that the proposed technique could adapt to larger problems while using the same amount of collected real-data. Studying the effectiveness and limitations of this technique with other classifiers and for other problem types is subject to a future extension of this work.

The COVID-19 pandemic hit in early 2020 impacted life in many aspects. Technology can play a critical role in identifying and tracking individuals who may be a source of infection \cite{filer2020test}. In a medical facility, staff and patients can be traced within buildings, besides door checks, to contain infection. Our future directions will focus on integrating our research with Internet of Things (IoT) devices that allow users (e.g., medical staff and patients) to walk around while tracked in a near real-time fashion. Privacy and Security are two important aspects that need to be studied along with technology development.   

\section{ACKNOWLEDGMENT}
This work is funded through a 4-VA collaborative research grant between Virginia Tech and James Madison University (https://4-va.org) - Spring 2018. The authors would like to thank Mr. Shawn Carddock (Director of Perioperative Services at Sentara hospital at the time of our experiment) for his support, and Mr. Marwan Khalil  (Virginia  Tech)  who  contributed  to the data collection stage of this project.

\bibliographystyle{./IEEEtran}
\input{main.bbl}

\end{document}